\documentclass[11pt,a4paper]{article}
\RequirePackage{xcolor}
\usepackage[hyperref]{acl2021}
\usepackage{times}
\usepackage{latexsym}
\usepackage{multirow}
\usepackage{graphicx}
\usepackage{makecell}
\usepackage{tabularx}
\usepackage{lscape}
\usepackage{longtable}
\usepackage{color, colortbl}
\usepackage{booktabs}
\usepackage[normalem]{ulem}
\usepackage{pgfplots}
\pgfplotsset{compat=1.15}
\usepackage{float}
\useunder{\uline}{\ul}{}

\newcommand{\method}{\textsc{CronKGQA}}
\newcommand{\dataset}{\textsc{CronQuestions}}

\makeatletter
\g@addto@macro{\normalsize}{%
\setlength{\abovedisplayskip}{3pt plus1pt}%
\setlength{\abovedisplayshortskip}{3pt plus1pt}%
\setlength{\belowdisplayskip}{3pt plus1pt}%
\setlength{\belowdisplayshortskip}{3pt plus1pt}}
\makeatother

\RequirePackage[textsize=scriptsize,color=yellow!15,linecolor=red,disable]{todonotes}
\RequirePackage{enumitem}
\setlist[itemize]{nosep,leftmargin=*,labelwidth=0pt}
\setlist[enumerate]{nosep, leftmargin=*}
\setlist[description]{nosep,leftmargin=.8em}

\RequirePackage{amsmath,amsthm,bm,amssymb}
\usepackage{pifont}
\newcommand{\cmark}{\ding{51}}%
\newcommand{\xmark}{\ding{55}}%

\usepackage{microtype}

\aclfinalcopy 

\def\bs{\bm{u}_s}
\def\bo{\bm{u}_o}
\def\bt{\bm{u}_t}
\def\rSO{\bm{v}_r^\text{SO}}
\def\rST{\bm{v}_r^\text{ST}}
\def\rOT{\bm{v}_r^\text{OT}}

\def\ztitle{Question Answering Over Temporal Knowledge Graphs}
\title{\ztitle}

\author{Apoorv Saxena \\
  Indian Institute of Science \\ Bangalore \\
  {\small \texttt{apoorvsaxena@iisc.ac.in}} \\\And
  Soumen Chakrabarti \\
  Indian Institute of Technology \\ Bombay \\
  {\small \texttt{soumen@cse.iitb.ac.in}} \\ \And
  Partha Talukdar \\
  Google Research \\ India \\
  {\small \texttt{partha@google.com}} \\
  }

\date{}

\begin{document}
\maketitle

\begin{abstract}
Temporal Knowledge Graphs (Temporal KGs) extend regular Knowledge Graphs by providing temporal scopes (e.g., start and end times) on each edge in the KG. While Question Answering over KG (KGQA) has received some attention from the research community, QA over Temporal KGs (Temporal KGQA) is a relatively unexplored area. Lack of broad-coverage datasets has been another factor limiting progress in this area. We address this challenge by presenting \dataset{}, the largest known Temporal KGQA dataset, clearly stratified into buckets of structural complexity. \dataset{} expands the only known previous dataset by a factor of 340$\times$. We find that various state-of-the-art KGQA methods fall far short of the desired performance on this new dataset.  In response, we also propose \method, a transformer-based solution that exploits recent advances in Temporal KG embeddings, and achieves performance superior to all baselines, with an increase of 120\% in accuracy over the next best performing method. Through extensive experiments, we give detailed insights into the workings of \method, as well as situations where significant further improvements appear possible. In addition to the dataset, we have released our code as well.
\end{abstract}

\begin{table*}[ht!]
\resizebox{\textwidth}{!}{%
\begin{tabular}{l|l|c|c|c|l|l}
\multicolumn{1}{c|}{}                                   & \multicolumn{1}{c|}{}                              &                                                                                      & \multicolumn{3}{c|}{\textbf{Question Types}}                                             & \multicolumn{1}{c}{}                                        \\ \cline{4-6}
\multicolumn{1}{c|}{\multirow{-2}{*}{\textbf{Dataset}}} & \multicolumn{1}{c|}{\multirow{-2}{*}{\textbf{KG}}} & \multirow{-2}{*}{\textbf{\begin{tabular}[c]{@{}c@{}}Temporal \\ facts\end{tabular}}} & \textbf{Multi-Entity} & \textbf{Multi-Relation} & \multicolumn{1}{c|}{\textbf{Temporal}} & \multicolumn{1}{c}{\multirow{-2}{*}{\textbf{\# questions}}} \\ \hline
SimpleQuestions                                         & FreeBase                                           & \xmark                                                                & \xmark & \xmark   & 0\%                                    & 108k                                                        \\
\rowcolor[HTML]{EFEFEF} 
MetaQA                                                  & MetaQA KG                                          & \xmark                                                                & \xmark & \cmark   & 0\%                                    & 400k                                                        \\
WebQuestions                                            & FreeBase                                           & \xmark                                                                & \xmark & \cmark   & \textless{}16\%                        & 5,810                                                       \\
\rowcolor[HTML]{EFEFEF} 
ComplexWebQuestions                                     & FreeBase                                           & \xmark                                                                & \cmark & \cmark   & -                                      & 35k                                                         \\
TempQuestions                                           & FreeBase                                           & \xmark                                                                & \cmark & \cmark   & 100\%                                  & 1,271                                                       \\
\rowcolor[HTML]{EFEFEF} 
\dataset{} (ours)                                                    & WikiData                                           & \cmark                                                                & \cmark & \cmark   & 100\%                                  & 410k                                                       
\end{tabular}%
}
\caption{KGQA dataset comparison. Statistics about percentage of temporal questions for WebQuestions are taken from \citet{tempquestions2018}. We do not have an explicit number of temporal questions for ComplexWebQuestions, but since it is constructed automatically using questions from WebQuestions, we expect the percentage to be similar to WebQuestions (16\%). Please refer to Section \ref{sec:temporal-qa-datasets} for details.}
\label{tab:tick-table}
\end{table*}
\section{Introduction}
\label{sec:Intro}


Temporal Knowledge Graphs (Temporal KGs) are multi-relational graph where each edge is associated with a time duration. This is in contrast to a regular KG where no time annotation is present. For example, a regular KG may contain a fact such as (\textit{Barack Obama}, \textit{held position}, \textit{President of USA}), while a temporal KG would contain the start and end time as well --- (\textit{Barack Obama}, \textit{held position}, \textit{President of USA}, \textit{2008}, \textit{2016}). Edges may be associated with a set of non-contiguous time intervals as well. These temporal scopes on facts can be either automatically estimated \cite{talukdar2012coupled} or user contributed. Several such Temporal KGs have been proposed in the literature, where the focus is on KG completion 
(\citealt{dasgupta-etal-2018-hyte}; \citealt{garcia-duran-etal-2018-learning}; \citealt{leetaru2013gdelt}; \citealt{lacroix2020tntcomplex}; \citealt{jain-etal-2020-temporal}).

The task of Knowledge Graph Question Answering (KGQA) is to answer natural language questions using a KG as the knowledge base. This is in contrast to reading comprehension-based question answering, where typically the question is accompanied by a context (e.g., text passage) and the answer is either one of multiple choices \citep{rajpurkar2016squad} or a piece of text from the context \citep{yang2018hotpotqa}. In KGQA, the answer is usually an entity (node) in the KG, and the reasoning required to answer questions is either single-fact based \citep{bordes2015large}, multi-hop (\citealt{yih-etal-2015-semantic}, \citealt{zhang2017variational}) or conjunction/comparison based reasoning \citep{talmor-berant-2018-web}. Temporal KGQA takes this a step further where:
\begin{enumerate}
    \item The underlying KG is a Temporal KG.
    \item The answer is either an entity or time duration.
    \item Complex temporal reasoning might be needed.
\end{enumerate}
\vspace{1mm}

KG Embeddings are low-dimensional dense vector representations of entities and relations in a KG. Several methods have been proposed in the literature to embed KGs (\citealt{bordes2013translating}, \citealt{trouillon2016complex}, \citealt{vashishth2020interacte}). These embeddings were originally proposed for the task of KG completion i.e.,  predicting missing edges in the KG, since most real world KGs are incomplete. Recently, however, they have also been applied to the task of KGQA where they have been shown to increase performance the settings of both of complete and incomplete KGs (\citealt{saxena-etal-2020-improving}; \citealt{sun2020faithful}). 

Temporal KG embeddings are another upcoming area where entities, relations and timestamps in a temporal KG are embedded in a low-dimensional vector space (\citealt{dasgupta-etal-2018-hyte}, \citealt{lacroix2020tntcomplex}, \citealt{jain-etal-2020-temporal}, \citealt{goel2019diachronic}). Here too, the main application so far has been temporal KG completion. In our work, we investigate whether temporal KG Embeddings can be applied to the task of Temporal KGQA, and how they fare compared to non-temporal embeddings or off-the-shelf methods without any KG Embeddings.


In this paper we propose \dataset{}, a new dataset for Temporal KGQA. \dataset{} consists of both a temporal KG and accompanying natural language questions. There were three main guiding principles while creating this dataset:
\begin{enumerate}
    \item The associated KG must provide temporal annotations.
    \item Questions must involve an element of temporal reasoning.
    \item The number of labeled instances must be large enough that it can be used for training models, rather than for evaluation alone.
\end{enumerate}
Guided by the above principles, we present a dataset consisting of a Temporal KG with 125k entities and 328k facts, along with a set of 410k natural language questions that require temporal reasoning.

On this new dataset, we apply approaches based on deep language models (LM) alone, such as T5 \citep{raffel2020exploring}, BERT \citep{devlin2019bert}, and KnowBERT \citep{peters2019knowledge}, and also hybrid LM+KG embedding approaches, such as Entities-as-Experts \citep{fevry2020entexperts} and EmbedKGQA \citep{saxena-etal-2020-improving}. We find that these baselines are not suited to temporal reasoning.  In response, we propose \method{}, an enhancement of EmbedKGQA, which outperforms baselines across all question types. \method{} achieves very high accuracy on simple temporal reasoning questions, but falls short when it comes to questions requiring more complex reasoning. Thus, although we get promising early results, \dataset{} leaves ample scope to improve complex Temporal KGQA.
\ifaclfinal 
    Our source code along with the \dataset{} dataset can be found at \url{https://github.com/apoorvumang/CronKGQA}.
\else
    Our dataset \dataset{} along with source code to reproduce results is available in the supplementary material. 
\fi

\begin{table*}[ht!]
\resizebox{\textwidth}{!}{%
\begin{tabular}{l|l|l}
\multicolumn{1}{c|}{\textbf{Reasoning}} & \multicolumn{1}{c|}{\textbf{Example Template}}     & \multicolumn{1}{c}{\textbf{Example Question}}         \\ \hline
Simple time                             & When did \{head\} hold the position of \{tail\}    & \textit{When did Obama hold the position of President of USA } \\
Simple entity                           & Which award did \{head\} receive in \{time\}       & \textit{Which award did Brad Pitt receive in 2001 }            \\
Before/After                            & Who was the \{tail\} \{type\} \{head\}             & \textit{Who was the President of USA before Obama }            \\
First/Last                              & When did \{head\} play their \{adj\} game          & \textit{When did Messi play their first game }                 \\
Time join                               & Who held the position of \{tail\} during \{event\} & \textit{Who held the position of President of USA during WWII}
\end{tabular}%
}
\caption{Example questions for different types of temporal reasoning. \{head\}, \{tail\} and \{time\} correspond to entities/timestamps in facts of the form (head, relation, tail, timestamp). \{event\} corresponds to entities in event facts eg. \textit{WWII}. \{type\} can be one of before/after and \{adj\} can be one of first/last. Please refer to Section \ref{sec:making-temporal-questions} for details.}
\label{tab:template}
\end{table*}

\section{Related work}

\subsection{Temporal QA data sets}
\label{sec:temporal-qa-datasets}

There have been several KGQA datasets proposed in the literature (Table~\ref{tab:tick-table}). In Simple\-Questions \citep{bordes2015large} one needs to extract just a single fact from the KG to answer a question. MetaQA \citep{zhang2017variational} and Web\-Questions\-SP \citep{yih-etal-2015-semantic} require multi-hop reasoning, where one must traverse over multiple edges in the KG to reach the answer. Complex\-Web\-Questions \citep{talmor-berant-2018-web} contains both multi-hop and conjunction/comparison type questions. However, none of these are aimed at temporal reasoning, and the KG they are based on is non-temporal.

Temporal QA datasets have mostly been studied in the area of reading comprehension. One such dataset is TORQUE \citep{ning2020torque}, where the system is given a question along with some context (a text passage) and is asked to answer a multiple choice question with five choices. This is in contrast to KGQA, where there is no context, and the answer is one of potentially hundreds of thousands of entities.

TempQuestions \citep{tempquestions2018} is a KGQA dataset specifically aimed at temporal QA. It consists of a subset of questions from Web\-Questions, Free917 \citep{cai-yates-2013-large} and Complex\-Questions \citep{bao-etal-2016-constraint} that are temporal in nature. They gave a definition for ``temporal question" and used certain trigger words (for example `before', `after') along with other constraints to filter out questions from these datasets that fell under this definition. However, this dataset contains only 1271 questions --- useful only for evaluation --- and the KG on which it is based (a subset of FreeBase \citep{freebase2008}) is not a temporal KG. Another drawback is that FreeBase has not been under active development since 2015, therefore some information stored in it is outdated and this is a potential source of inaccuracy.


\subsection{Temporal QA algorithms}

To the best of our knowledge, recent KGQA algorithms (\citealt{miller2016kvmnet}; \citealt{sun2019pullnet}; \citealt{cohen2020scalable}; \citealt{sun2020faithful}) work with \textit{non-temporal KGs}, i.e., KGs containing facts of the form (subject, relation, object). Extending these to \textit{temporal KGs} containing facts of the form (subject, relation, object, start time, end time) is a non-trivial task. TEQUILA \citep{Jia_2018} is one method aimed specifically at temporal KGQA. TEQUILA decomposes and rewrites the question into non-temporal sub-questions and temporal constraints. Answers to sub-questions are then retrieved using any KGQA engine. Finally, TEQUILA uses constraint reasoning on temporal intervals to compute final answers to the full question. A major drawback of this approach is the use of pre-specified templates for decomposition, as well as the assumption of having temporal constraints on entities. Also, since it is made for non-temporal KGs, there is no direct way of applying it to temporal KGs where facts are temporally scoped.

\section{\dataset{}: The new Temporal KGQA dataset}

\dataset{}, our Temporal KGQA dataset consists of two parts: a KG with temporal annotations, and a set of natural language questions requiring temporal reasoning.

\subsection{Temporal KG}

To prepare our temporal KG, we started by taking all facts with temporal annotations from the WikiData subset proposed by \citet{lacroix2020tntcomplex}. We removed some instances of the predicate ``\textit{member of sports team}'' in order to balance out the KG since this predicate constituted over 50 percent of the facts. Timestamps were discretized to years. This resulted in a KG with 323k facts, 125k entities and 203 relations.

However, this filtering of facts misses out on important world events. For example, the KG subset created using the aforementioned technique contains the entity \textit{World War II} but no associated fact that tells us when \textit{World War II} started or ended. This knowledge is needed to answer questions such as ``\textit{Who was the President of the USA during World War II?}.''  To overcome this shortcoming, we first extracted entities from WikiData that have a ``start time'' and ``end time'' annotation. From this set, we then removed entities which were game shows, movies or television series (since these are not important world events, but do have a start and end time annotation), and then removed entities with less than 50 associated facts. This final set of entitities was then added as facts in the format (\textit{WWII, significant event, occurred, 1939, 1945)}. The final Temporal KG consisted of 328k facts out of which 5k are event-facts.

\begin{table}[t!]
\resizebox{\columnwidth}{!}{%
\begin{tabular}{l|l}
Template            & \textit{When did \{head\} play in \{tail\}} \\ \hline
Seed Qn             & \textit{When did \textbf{Messi} play in \textbf{FC Barcelona}}   \\ \hline
 \makecell{Human \\ Paraphrases}   & 
 
 \makecell[l]{
 \textit{When was \textbf{Messi} playing in \textbf{FC Barcelona}} \\
 \textit{Which years did \textbf{Messi} play in \textbf{FC Barcelona}} \\
 \textit{When did \textbf{FC Barcelona} have \textbf{Messi} in their team} \\
 \textit{What time did \textbf{Messi} play in \textbf{FC Barcelona}} \\
 }    \\ \hline
\makecell{Machine \\ Paraphrases} & 
\makecell[l]{
\textit{When did \textbf{Messi} play for \textbf{FC Barcelona}} \\
\textit{When did \textbf{Messi} play at \textbf{FC Barcelona}} \\
\textit{When has \textbf{Messi} played at \textbf{FC Barcelona}}
}
\end{tabular}
}
\caption{Slot-filled paraphrases generated by humans and machine. Please refer to Section~\ref{sec:making-temporal-questions} for details.}
\label{tab:pp}
\end{table}


\begin{table}[t!]
\resizebox{\columnwidth}{!}{%
\begin{tabular}{l|rrr}
\hline
               & \multicolumn{1}{l}{\textbf{Train}} & \multicolumn{1}{l}{\textbf{Dev}} & \multicolumn{1}{l}{\textbf{Test}} \\ \hline
Simple Entity  & 90,651                             & 7,745                            & 7,812                             \\
Simple Time    & 61,471                             & 5,197                            & 5,046                             \\
Before/After   & 23,869                             & 1,982                            & 2,151                             \\
First/Last     & 118,556                            & 11,198                           & 11,159                            \\
Time Join      & 55,453                             & 3,878                            & 3,832                             \\ \hline
Entity Answer  & 225,672                            & 19,362                           & 19,524                            \\
Time Answer    & 124,328                            & 10,638                           & 10,476                            \\ \hline
\textbf{Total} & \multicolumn{1}{l}{350,000}        & \multicolumn{1}{l}{30,000}       & \multicolumn{1}{l}{30,000}        \\ \hline
\end{tabular}
}
\caption{Number of questions in our dataset across different types of reasoning required and different answer types. Please refer to Section~\ref{sec:question-categorization} for details.}
\label{tab:dataset-stats-questions}
\end{table}

\subsection{Temporal Questions}
\label{sec:making-temporal-questions}

To generate the QA dataset, we started with a set of templates for temporal reasoning. These were made using the five most frequent relations from our WikiData subset, namely 
\begin{itemize}
    \item \textit{member of sports team}
    \item \textit{position held}
    \item \textit{award received}
    \item \textit{spouse}
    \item \textit{employer}
\end{itemize}

This resulted in 30 unique seed templates over five relations and five different reasoning structures (please see Table~\ref{tab:template} for some examples). Each of these templates has a corresponding procedure that could be executed over the temporal KG to extract all possible answers for that template. However, similar to \citet{zhang2017variational}, we chose not to make this procedure a part of the dataset, to remove unwelcome dependence of QA systems on such formal candidate collection methods. This also allows easy augmentation of the dataset, since only question-answer pairs are needed.

In the same spirit as ComplexWebQuestions, we then asked human annotators to paraphrase these templates in order to generate more linguistic diversity. Annotators were given slot-filled templates with dummy entities and times, and asked to rephrase the question such that the dummy entities/times were present in the paraphrase and the question meaning did not change. This resulted in 246 unique templates.

We then used the monolingual paraphraser developed by \citet{N18-1007} to automatically generate paraphrases using these 246 templates. After verifying their correctness through annotators, we ended up with 654 templates. These templates were then filled using entity aliases from WikiData to generate 410k unique question-answer pairs. 

Finally, while splitting the data into train/test folds, we ensured that 
\begin{enumerate}
    \item Paraphrases of train questions are not present in test questions.
    \item There is no entity overlap between test questions and train questions. Event overlap is allowed.
\end{enumerate}
The second requirement implies that, if the question ``\textit{Who was president before Obama}'' is present in the train set, the test set cannot contain any question that mentions the entity `\textit{Obama}'.
While this policy may appear like an overabundance of caution, it ensures that models are doing temporal reasoning rather than guessing from entities seen during training. \citet{lewis2020question} noticed an issue in WebQuestions where they found that almost 30\% of test questions overlapped with training questions. The issue has been seen in the MetaQA dataset as well, where there is significant overlap between test/train entities and test/train question paraphrases, leading to suspiciously high performance on baseline methods even with partial KG data \citep{saxena-etal-2020-improving}, which suggests that models that apparently perform well are not necessarily performing the desired reasoning over the KG.

A drawback of our data creation protocol is that question/answer pairs are generated automatically.  Therefore, the question distribution is artificial from a semantic perspective.  (Complex\-Web\-Questions has a similar limitation.)  However, since developing models that are capable of temporal reasoning is an important direction for natural language understanding, we feel that our dataset provides an opportunity to both train and evaluate KGQA models because of its large size, notwithstanding its lower-than-natural linguistic variety. In Section~\ref{sec:training-data-size-effect}, we show the effect that training data size has on model performance.

Summarizing, each of our examples contains
\begin{enumerate}
    \item A paraphrased natural language question.
    \item A set of entities/times in the question.
    \item A set of `gold' answers (entity or time).
\end{enumerate}

The entities are specified as WikiData IDs (e.g., \textit{Q219237}), and times are years (e.g., \textit{1991}). We include the set of entities/times in the test questions as well since similar to other KGQA datasets (MetaQA, WebQuestions, ComplexWebQuestions) and methods that use these datasets (PullNet, EmQL), entity linking is considered as a separate problem and complete entity linking is assumed. We also include the seed template and head/tail/time annotation in the train fold, but omit these from the test fold.
\begin{figure*}
  \centering
  \includegraphics[width=\textwidth, trim={0.6cm 0 0.4cm 0},clip ]{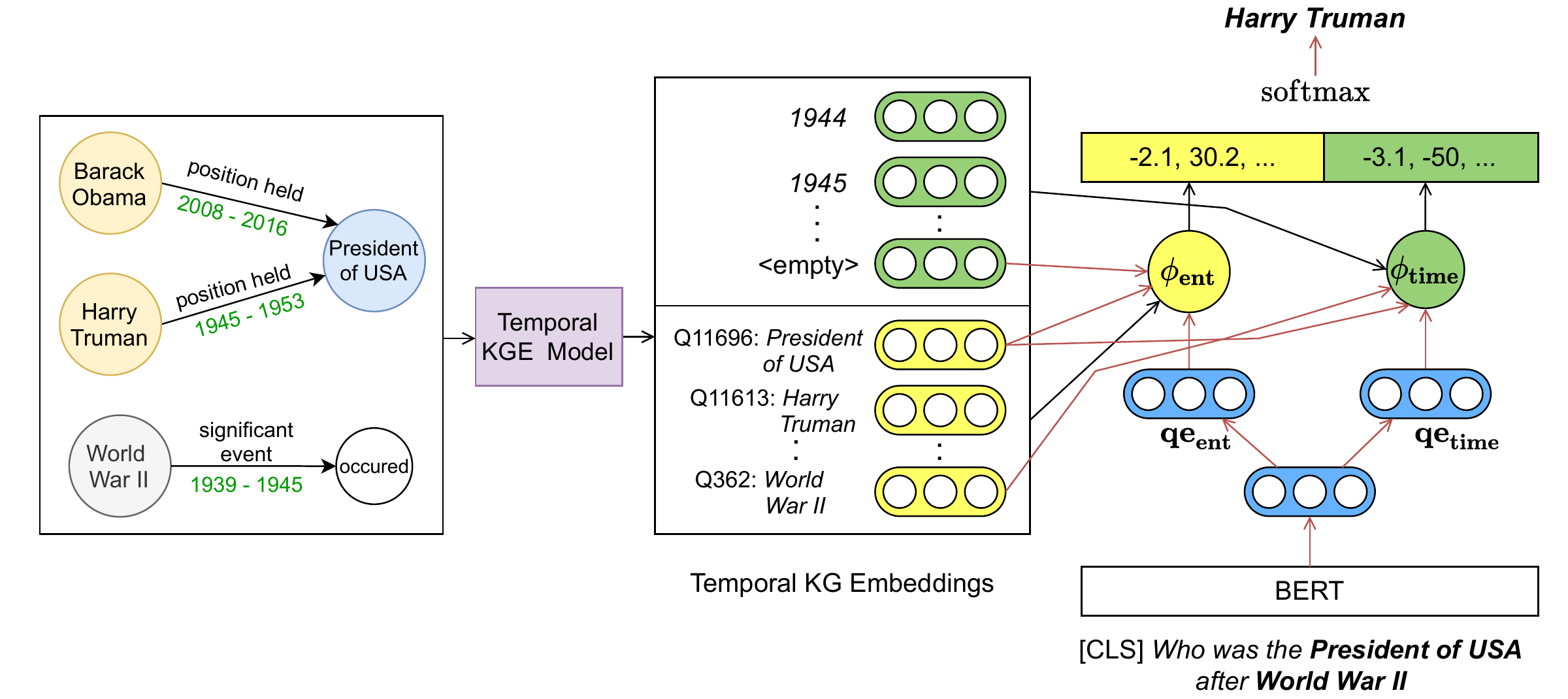}
  \caption{The \method{} method. (i)~A temporal KG embedding model (Section \ref{sec:temporal-kg-embeddings}) is used to generate embeddings for each timestamp and entity in the temporal knowledge graph (ii)~BERT is used to get two question embeddings: $\bm{qe}_{ent}$ and $\bm{qe}_{time}$. (iii)~Embeddings of entity/time mentions in the question are combined with question embeddings using equations \ref{eqn:ent-score} and \ref{eqn:time-score} to get score vectors for entity and time prediction. (iv)~Score vectors are concatenated and softmax is used get answer probabilities. Please refer to Section \ref{sec:TembedKGQA} for details.}
  \label{fig:our-model}
\end{figure*}
\subsubsection{Question Categorization}
\label{sec:question-categorization}

In order to aid analysis, we categorize questions into ``simple reasoning'' and ``complex reasoning'' questions (please refer to Table~\ref{tab:dataset-stats-questions} for the distribution statistics). 
\begin{description}
\item[Simple reasoning:] These questions require a single fact to answer, where the answer can be either an entity or a time instance. For example the question \textit{``Who was the President of the United States in 2008?"} requires a single fact to answer the question, namely (\textit{Barack Obama}, \textit{held position}, \textit{President of USA}, \textit{2008}, \textit{2016})
\item[Complex reasoning:] These questions require multiple facts to answer and can be more varied. For example \textit{``Who was the first President of the United States?"}  This requires reasoning over multiple facts pertaining to the entity \textit{``President of the United States''}. In our dataset, all questions that are not ``simple reasoning'' questions are considered complex questions. These are further categorized into the types ``before/after`', ``first/last'' and ``time join'' --- please refer Table~\ref{tab:template} for examples of these questions.

\end{description}

\section{Temporal KG Embeddings}
\label{sec:temporal-kg-embeddings}
We investigate how we can use KG embeddings, both temporal and non-temporal, along with pre-trained language models to perform temporal KGQA.  We will first briefly describe the specific KG embedding models we use, and then go on to show how we use them in our QA models.  In all cases, the scores are turned into suitable losses with regard to positive and negative tuples in an incomplete KG, and these losses minimized to train the entity, time and relation representations.

\subsection{ComplEx}
\label{sec:complex}
ComplEx \citep{trouillon2016complex} represents each entity $e$ as a complex vector $\bm{u}_e\in\mathbb{C}^D$.  Each relation $r$ is represented as a complex vector $\bm{v}_r\in\mathbb{C}^D$ as well. The score $\phi$ of a claimed fact $(s,r,o)$ is
\begin{align}
    \phi(s,r,o) &= \Re(\langle \bm{u}_s, \bm{v}_r, \bm{u}_o^\star \rangle) \notag \\
&= \Re\big(\textstyle\sum_{d=1}^D \bm{u}_s[d] \bm{v}_r[d] \bm{u}_o[d]^\star \big)
\label{eqn:complex}
\end{align}
where $\Re(\cdot)$ denotes the real part and $c^\star$ is the complex conjugate.  Despite further developments, ComplEx, along with refined training protocols \citep{lacroix2018canonical} remains among the strongest KB embedding approaches \citep{Ruffinelli2020You}.

\subsection{TComplEx, TNTComplEx}
\label{sec:tcomplex}

\citet{lacroix2020tntcomplex} took an early step to extend ComplEx with time. Each timestamp $t$ is also represented as a complex vector $\bm{w}_t\in\mathbb{C}^D$. For a claimed fact $(s, r, o, t)$, their TComplEx scoring function is
\begin{equation}
    \begin{aligned}
        \phi(s,r,o, t) &= \Re(\langle \bm{u}_s, \bm{v}_r, \bm{u}_o^\star, \bm{w}_t \rangle)
    \end{aligned}
    \label{eqn:tcomplex}
\end{equation}
Their TNTComplEx scoring function uses two representations of relations $r$: $\bm{v}^\text{T}_r$, which is sensitive to time, and $\bm{v}_r$, which is not.  The scoring function is the sum of a time-sensitive and a time-insensitive part: $\Re(\langle \bm{u}_s, \bm{v}^\text{T}_r, \bm{u}_o^\star, \bm{w}_t \rangle + \langle \bm{u}_s, \bm{v}_r, \bm{u}_o^\star, \mathbf{1} \rangle)$.

\subsection{TimePlex}

TimePlex \citep{jain-etal-2020-temporal} augmented ComplEx with embeddings $\bm{u}_t \in \mathbb{C}^D$ for discretized time instants~$t$.  To incorporate time, TimePlex uses three representations for each relation $r$, viz., $(\bm{v}^\text{SO}_r, \bm{v}^\text{ST}_r, \bm{v}^\text{OT}_r)$ and writes the base score of a tuple $(s, r, o, t)$ as 
\begin{multline}
    \phi(s, r, o, t) = \langle \bs, \rSO, \bo^\star\rangle + \alpha\,\langle \bs, \rST, \bt^\star\rangle \\
    + \beta\, \langle \bo, \rOT, \bt^\star\rangle + \gamma\,\langle \bs, \bo, \bt^\star\rangle,
    \label{eqn:timeplex}    
\end{multline}
where $\alpha,\beta,\gamma$ are hyperparameters.  

\section{\method{}: Our proposed method}
\label{sec:TembedKGQA}


We start with a temporal KG, apply a time-agnostic or time-sensitive KG embedding algorithm (ComplEx, TComplEx, or TimePlex) to it, and obtain entity, relation, and timestamp embeddings for the temporal KG. We will use the following notation.
\begin{itemize}
    \item $\mathcal{E}$ is the matrix of entity embeddings
    \item $\mathcal{T}$ is the matrix of timestamp embeddings
    \item $\mathcal{E.T}$ is the concatenation of $\mathcal{E}$ and $\mathcal{T}$ matrices. This is used for scoring answers, since the answer can be either an entity or timestamp.
\end{itemize}
In case entity/timestamp embeddings are complex valued vectors in $\mathbb{C}^D$, we expand them to real valued vectors of size $2D$, where the first half is the real part and the second half is the complex part of the original vector.

We first apply EmbedKGQA \citep{saxena-etal-2020-improving}  directly to the task of Temporal KGQA. In its original implementation, EmbedKGQA uses ComplEx (Section \ref{sec:complex}) embeddings and can only deal with non-temporal KGs and single entity questions. In order to apply it to \dataset{}, we set the first entity encountered in the question as the ``\textit{head entity}'' needed by EmbedKGQA. Along with this, we set the entity embedding matrix  $\mathcal{E}$ to be the ComplEx embedding of our KG entities, and initialize $\mathcal{T}$ to a random learnable matrix. EmbedKGQA then performs prediction over $\mathcal{E.T}$.


Next, we modify EmbedKGQA so that it can use temporal KG embeddings. We use TComplEx (Section \ref{sec:tcomplex}) for getting entity and timestamp embeddings. \method{} (Figure \ref{fig:our-model}) utilizes two scoring functions, one for predicting entity and one for predicting time. Using a pre-trained LM (BERT in our case) \method{} finds a question embedding $\bm{qe}$. This is then projected to get two embeddings, $\bm{qe}_{ent}$ and $\bm{qe}_{time}$, which are question embeddings for entity and time prediction respectively.

\begin{description}
\item[Entity scoring function:] We extract a subject entity $s$ and a timestamp $t$ from the question. If either is missing, we use a dummy entity/time. Then, using the scoring function $\phi(s, r, o, t)$ from equation~\ref{eqn:tcomplex}, we calculate a score for each entity $e \in \mathbf{E}$ as 
\begin{equation}
\label{eqn:ent-score}
\begin{aligned}
    \bm{\phi_{ent}}(e) = \Re(\langle \bm{u}_s, \bm{qe}_{ent}, \bm{u}_e^\star, \bm{w}_t \rangle)
\end{aligned}
\end{equation}
where $\mathbf{E}$ is the set of entities in the KG. This gives us a score for each entity being an answer. 

\item[Time scoring function:] Similarly, we extract a subject entity $s$ and object entity $o$ from the question, using dummy entities if none are present. Then, using~\ref{eqn:tcomplex}, we calculate a score for each timestamp $t \in \mathbf{T}$ as
\begin{equation}
\label{eqn:time-score}
\begin{aligned}
    \bm{\phi_{time}}(t) = \Re(\langle \bm{u}_s, \bm{qe}_{time}, \bm{u}_o^\star, \bm{w}_t \rangle)
\end{aligned}
\end{equation}

\end{description}
The scores for all entities and times are concatenated, and $\mathrm{softmax}$ is used to calculate answer probabilities over this combined score vector. The model is trained using cross entropy loss.

\begin{table*}[ht!]
\resizebox{\textwidth}{!}{%
\begin{tabular}{l|r|rr|rr|rrrrr}
\multicolumn{1}{c|}{\multirow{3}{*}{\textbf{Model}}} & \multicolumn{5}{c|}{\textbf{Hits@1}}                                                                                                                                                                              & \multicolumn{5}{c}{\textbf{Hits@10}}                                                                                                                                                                             \\ \cline{2-11} 
\multicolumn{1}{c|}{}                                & \multicolumn{1}{c|}{\multirow{2}{*}{\textbf{Overall}}} & \multicolumn{2}{c|}{\textbf{Question Type}}                                  & \multicolumn{2}{c|}{\textbf{Answer Type}}                                 & \multicolumn{1}{c|}{\multirow{2}{*}{\textbf{Overall}}} & \multicolumn{2}{c|}{\textbf{Question Type}}                                  & \multicolumn{2}{c}{\textbf{Answer Type}}                                \\ \cline{3-6} \cline{8-11} 
\multicolumn{1}{c|}{}                                & \multicolumn{1}{c|}{}                                  & \multicolumn{1}{c|}{\textbf{Complex}} & \multicolumn{1}{c|}{\textbf{Simple}} & \multicolumn{1}{c|}{\textbf{Entity}} & \multicolumn{1}{c|}{\textbf{Time}} & \multicolumn{1}{c|}{}                                  & \multicolumn{1}{c|}{\textbf{Complex}} & \multicolumn{1}{c|}{\textbf{Simple}} & \multicolumn{1}{c|}{\textbf{Entity}} & \multicolumn{1}{c}{\textbf{Time}} \\ \hline
BERT                                                 & 0.071                                                  & 0.086                                 & 0.052                                & 0.077                                & 0.06                              & \multicolumn{1}{r|}{0.213}                             & 0.205                                 & \multicolumn{1}{r|}{0.225}           & 0.192                                 & 0.253                             \\
RoBERTa                                              & \multicolumn{1}{r|}{0.07}            & 0.086                                       & 0.05                                & 0.082                                      & \multicolumn{1}{r|}{0.048}          & \multicolumn{1}{r|}{0.202}            & 0.192                                       & \multicolumn{1}{r|}{0.215}           & 0.186                                     & 0.231                             \\
KnowBERT                                             & 0.07                                                  & 0.083                                 & 0.051                                & 0.081                                & 0.048                              & \multicolumn{1}{r|}{0.201}                              & 0.189                                 & \multicolumn{1}{r|}{0.217}           & 0.185                                & 0.23                             \\
T5-3B                                                & 0.081                                                  & 0.073                                 & 0.091                                & 0.088                                & 0.067                              & \multicolumn{1}{r|}{-}                                 & -                                     & \multicolumn{1}{r|}{-}               & -                                    & -                                 \\ \hline
EmbedKGQA                                            & 0.288                                                   & 0.286                                  & 0.29                                & { 0.411}                          & 0.057                              & \multicolumn{1}{r|}{0.672}                             & 0.632                                 & \multicolumn{1}{r|}{0.725}           & {0.85}                          & 0.341                             \\
T-EaE-add                                              & 0.278                                                   & 0.257                                  & 0.306                                & 0.313                                & 0.213                               & \multicolumn{1}{r|}{0.663}                             & 0.614                                 & \multicolumn{1}{r|}{0.729}           & 0.662                                & 0.665                             \\
T-EaE-replace                                          & { 0.288}                                            & { 0.257}                           & { 0.329}                          & 0.318                                & { 0.231}                        & \multicolumn{1}{r|}{{ 0.678}}                       & { 0.623}                           & \multicolumn{1}{r|}{{ 0.753}}     & 0.668                                & { 0.698}                       \\
\method{}                                          & \textbf{0.647}                                         & \textbf{0.392}                        & \textbf{0.987}                       & \textbf{0.699}                       & \textbf{0.549}                     & \multicolumn{1}{r|}{\textbf{0.884}}                    & \textbf{0.802}                        & \multicolumn{1}{r|}{\textbf{0.992}}  & \textbf{0.898}                       & \textbf{0.857}                   
\end{tabular}%
}
\caption{Performance of baselines and our methods on the \dataset{} dataset. Methods above the midrule do not use any KG embeddings, while the ones below use either temporal or non-temporal KG embeddings. Hits@10 are not available for T5-3B since it is a text-to-text model and makes a single prediction. Please refer to Section \ref{sec:main-results} for details.}
\label{tab:main-results}
\end{table*}




\section{Experiments and diagnostics}
\label{sec:Expt}
In this section, we aim to answer the following questions:
\begin{enumerate}
    \item How do baselines and \method{} perform on the \dataset{} task? (Section~\ref{sec:main-results}.)
    \item Do some methods perform better than others on specific reasoning tasks? (Section~\ref{sec:perf_across_qn_type}.)
    \item How much does the training dataset size (number of questions) affect the performance of a model? (Section~\ref{sec:training-data-size-effect}.)
    \item Do temporal KG embeddings confer any advantage over non-temporal KG embeddings? (Section~\ref{sec:temp_vs_nontemp}.)
\end{enumerate}

\subsection{Other methods compared}

It has been shown by \citet{petroni2019language} and \citet{raffel2020exploring} that large LMs, such as BERT and its variants, capture real world knowledge (collected from their massive, encyclopedic training corpus) and can directly be applied to tasks such as QA. In these baselines, we do not specifically feed our version of the temporal KG to the model --- we instead expect the model to have the real world knowledge to compute the answer. 

\begin{description}
\item[BERT:] We experiment with BERT, RoBERTa \citep{liu2019roberta} and Know\-BERT \citep{peters2019knowledge} which is a variant of BERT where information from knowledge bases such as WikiData and WordNet has been injected into BERT. 
We add a prediction head on top of the [CLS] token of the final layer and do a $\mathrm{softmax}$ over it to predict the answer probabilities.

\item[T5:] In order to apply T5 \citep{raffel2020exploring} to temporal QA, we transform each question in our dataset to the form `\textit{temporal question: }$\langle \mathrm{question} \rangle$\textit{?}'. For evaluation there are two cases:
\begin{enumerate}
    \item Time answer: We do exact string matching between T5 output and correct answer.
    \item Entity answer: We compare the system output to the aliases of all entities in the KG. The entity having an alias with the smallest edit distance \citep{levenshtein1966binary} to the predicted text output is taken as the predicted entity.
\end{enumerate}

\item[Entities as experts:]\citet{fevry2020entexperts} proposed EaE, a model which aims to integrate entity knowledge into a transformer-based language model.
For temporal KGQA on \dataset{}, we assume that all grounded entity and time mention spans are marked in the question\footnote{This assumption can be removed by using EaE's early transformer stages as NE spotters and disambiguators.}. We will refer to this model as \textbf{T-EaE-add}. We try another variant of EaE, \textbf{T-EaE-replace}, where instead of adding the entity/time and BERT token embeddings, we replace the BERT embeddings with the entity/time embeddings for entity/time mentions.\footnote{Appendix~\ref{sec:eae_appendix} gives details of our EaE implementation.}

\end{description}

\subsection{Main results}
\label{sec:main-results}
Table~\ref{tab:main-results} shows the results of various methods on our dataset. We see that methods based on large pre-trained LMs alone (BERT, RoBERTa, T5), as well as KnowBERT, perform significantly worse than methods that are augmented with KG embeddings (temporal or non-temporal). This is probably because having KG embeddings specific to our temporal KG helps the model to focus on those entities/timestamps. In our experiments, BERT performs slightly better than KnowBERT, even though KnowBERT has entity knowledge in its parameters. T5-3B performs the best among the LMs we tested, possibly because of the large number of parameters and pre-training.

Even among methods that use KG embeddings, \method{} performs the best on all metrics, followed by T-EaE-replace. Since EmbedKGQA has non-temporal embeddings, its performance on questions where the answer is a time is very low --- comparable to BERT --- which is the LM used in our EmbedKGQA implementation.


Another interesting thing to note is the performance on simple reasoning questions. \method{} far outperforms baselines for simple questions, achieving close to 0.99 hits@1, which is much lower for T-EaE (0.329). We believe there might be a few reasons that contribute to this:
\begin{enumerate}
    \item There is the \textit{inductive bias} of combining embeddings using TComplEx scoring function in \method{}, which is the same one used in creating the entity and time embeddings, thus making the simple questions straightforward to answer. However, not relying on a scoring function means that T-EaE can be extended to any KG embedding, whereas \method{} cannot. 
    \item Another contributing reason could be that there are fewer parameters to be trained in \method{} while a 6-layer Transformer encoder needs to be trained from scratch in T-EaE. Transformers typically require large amounts of varied data to train successfully.
\end{enumerate}

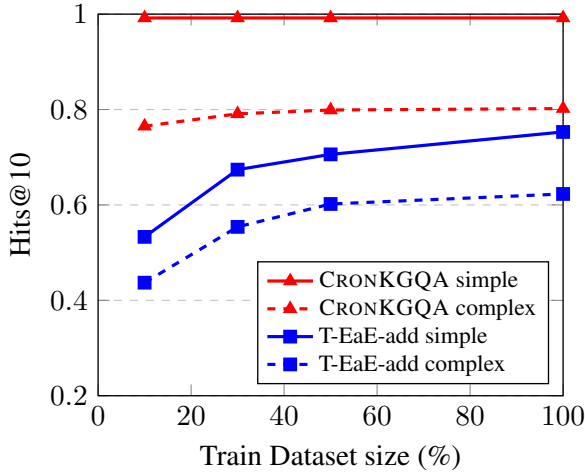
\begin{figure}[t!]
    \centering
   \begin{tikzpicture}
\begin{axis}[
    width=\columnwidth,
    line width=0.5,
    xlabel={Train Dataset size (\%)},
    ylabel={Hits@10},
    xmin=0, xmax=100,
    ymin=0.2, ymax=1,
    xtick={0,20,40,60,80,100},
    ytick={0.2,0.4,0.6,0.8,1.0},
    legend pos=south east,
    ymajorgrids=true,
    grid style=dashed,
    legend style={nodes={scale=0.8, transform shape}},
    legend cell align={left},
]

\addplot[
    red,very thick,solid,
    mark=triangle*,
    mark options={solid},
    ]
    coordinates {
    (10,0.992)(30,0.992)(50,0.992)(100,0.992)
    };

\addplot+[
    red,very thick,dashed,
    mark=triangle*,
    mark options={solid},
    ]
    coordinates {
    (10,0.765)(30,0.791)(50,0.799)(100,0.802)
    };

\addplot+[
    blue,very thick,solid,
    mark=square*,
    mark options={solid},
    ]
    coordinates {
    (10,0.533)(30,0.674)(50,0.706)(100,0.753)
    };

\addplot+[
    blue,very thick,dashed,
    mark=square*,
    mark options={solid},
    ]
    coordinates {
     (10,0.437)(30,0.554)(50,0.602)(100,0.623)
    };

    \addlegendentry{\method{} simple}
    \addlegendentry{\method{} complex}
    \addlegendentry{T-EaE-add simple}
    \addlegendentry{T-EaE-add complex}

\end{axis}
\end{tikzpicture}
    \caption{Model performance (hits@10) vs.\ training dataset size (percentage) for \method{} and T-EaE-add. Solid line is for simple reasoning and dashed line is for complex reasoning type questions. For each dataset size, models were trained until validation hits@10 did not increase for 10 epochs. Please refer to Section \ref{sec:training-data-size-effect} for details.}
    \label{fig:dataset-size}
\end{figure}

\subsection{Performance across question types}
\label{sec:perf_across_qn_type}

\begin{table*}[t!]
\centering
\resizebox{1.2\columnwidth}{!}{%
\begin{tabular}{l|rrrrr|r}
              & \multicolumn{1}{c}{\textbf{\begin{tabular}[c]{@{}c@{}}Before/\\ After\end{tabular}}} & \multicolumn{1}{c}{\textbf{\begin{tabular}[c]{@{}c@{}}First/\\ Last\end{tabular}}} & \multicolumn{1}{c}{\textbf{\begin{tabular}[c]{@{}c@{}}Time\\ Join\end{tabular}}} & \multicolumn{1}{c}{\textbf{\begin{tabular}[c]{@{}c@{}}Simple\\ Entity\end{tabular}}} & \multicolumn{1}{c|}{\textbf{\begin{tabular}[c]{@{}c@{}}Simple\\ Time\end{tabular}}} &
               \multicolumn{1}{c}{\textbf{\begin{tabular}[c]{@{}c@{}}All\end{tabular}}}\\ \hline
EmbedKGQA     & { 0.199}                                                                  & { 0.324}                                                               & 0.223                                                                   & { 0.421}                                                                 & 0.087     & 0.288                                                                \\
T-EaE-add     & 0.256                                                                       & 0.285                                                                     & 0.175                                                                   & 0.296                                                                       & 0.321           & 0.278                                                           \\
T-EaE-replace & 0.256                                                                       & 0.288                                                                     & { 0.168}                                                             & 0.318                                                                       & { 0.346}        & 0.288                                                       \\
\method{}     & \textbf{0.288}                                                              & \textbf{0.371}                                                            & \textbf{0.511}                                                          & \textbf{0.988}                                                              & \textbf{0.985}      & \textbf{0.647}                                                    
\end{tabular}%
}
\caption{Hits@1 for different reasoning type questions. `Simple Entity' and `Simple Time' correspond to simple question type in Table \ref{tab:main-results} while the others correspond to complex question type. Please refer to section \ref{sec:perf_across_qn_type} for more details.}
\label{tab:reasoning-comparison}
\end{table*}

Table~\ref{tab:reasoning-comparison} shows the performance of KG embedding based models across different types of reasoning. As stated above in Section \ref{sec:main-results}, \method{} performs very well on simple reasoning questions (simple entity, simple time). Among complex question types, all models (except EmbedKGQA) perform the best on time join questions (e.g., `\textit{Who played with Roberto Dinamite on the Brazil national football team}'). This is because such questions typically have multiple answers (such as all the players when \textit{Roberto Dinamite} was playing for Brazil), which makes it easier for the model to make a correct prediction. In the other two question types, the answer is always a single entity/time. Before/after questions seem most challenging for all methods, with the best method achieving only 0.288 hits@1.

\subsection{Effect of training dataset size}
\label{sec:training-data-size-effect}

Figure~\ref{fig:dataset-size} shows the effect of training dataset size on model performance. As we can see, for T-EaE-add, increasing the training dataset size from 10\% to 100\% steadily increases its performance for both simple and complex reasoning type questions. This effect is somewhat present in \method{} for complex reasoning, but not so for simple reasoning type questions. We hypothesize that this is because T-EaE has more trainable parameters --- it has a 6-layer transformer that needs to be trained from scratch --- in contrast to \method{} that needs to merely fine tune BERT and train some shallow projection layers. These results affirm our hypothesis that having a large, even if synthetic, dataset is useful for training temporal reasoning models.




\subsection{Temporal vs.\ non-temporal KG embeddings}
\label{sec:temp_vs_nontemp}

We conducted further experiments to study the effect of temporal vs.\ non-temporal KG embeddings. We replaced the temporal entity embeddings in T-EaE-replace with ComplEx embeddings, and treated timestamps as regular tokens (not associated with any entity/time mentions). \method{}-CX is the same as EmbedKGQA. The results can be seen in Table \ref{tab:complex-vs-tcomplex}. As we can see, for both \method{} and T-EaE-replace, using temporal KGE (TComplex) gives a significant boost in performance compared to non-temporal KGE (ComplEx). \method{} receives a much larger boost in performance compared to T-EaE-replace, probably because the scoring function has been modeled after TComplEx and not ComplEx, while there is no such embedding-specific engineering in T-EaE-replace. Another observation is that questions having temporal answers achieve very low accuracy (0.057 and 0.062 respectively) in both \method{}-CX and T-EaE-replace-CX, which is much lower than what these models achieve with TComplEx. This shows that having temporal KG embeddings is essential for achieving good performance for KG embedding-based methods.

\begin{table}[t!]
\resizebox{\columnwidth}{!}{%
\begin{tabular}{l|rr|rr}
\multicolumn{1}{c|}{\multirow{2}{*}{\textbf{\begin{tabular}[c]{@{}c@{}}Question\\ Type\end{tabular}}}} & \multicolumn{2}{c|}{\textbf{\method{}}}                                      & \multicolumn{2}{c}{\textbf{T-EaE-replace}}                                        \\ \cline{2-5} 
\multicolumn{1}{c|}{}                                                                                  & \multicolumn{1}{c|}{\textbf{CX}} & \multicolumn{1}{c|}{\textbf{TCX}} & \multicolumn{1}{c|}{\textbf{CX}} & \multicolumn{1}{c}{\textbf{TCX}} \\ \hline
Simple                                                                                                 & 0.29                                 & 0.987                                  & 0.248                                 & 0.329                                 \\
Complex                                                                                                & 0.286                                  & 0.392                                  & 0.247                                 & 0.257                                  \\ \hline
Entity Answer                                                                                          & 0.411                                 & 0.699                                  & 0.347                                 & 0.318                                 \\
Time Answer                                                                                            & 0.057                                 & 0.549                                  & 0.062                                 & 0.231                                  \\ \hline
Overall                                                                                                & 0.288                                  & 0.647                                  & 0.247                                 & 0.288                                 
\end{tabular}%
}
\caption{Hits@1 for \method{} and T-EaE-replace using ComplEx(CX) and TComplEx(TCX) KG embeddings. Please refer to Section \ref{sec:temp_vs_nontemp} for more details.}
\label{tab:complex-vs-tcomplex}
\end{table}



\section{Conclusion}

In this paper we introduce \dataset{}, a new dataset for Temporal Knowledge Graph Question Answering. While there exist some Temporal KGQA datasets, they are all based on non-temporal KGs (e.g., Freebase) and have relatively few questions. Our dataset consists of both a temporal KG as well as a large set of temporal questions requiring various structures of reasoning.  In order to develop such a large dataset, we used a  synthetic generation procedure, leading to a question distribution that is artificial from a semantic perspective. However, having a large dataset provides an opportunity to train models, rather than just evaluate them.  We experimentally show that increasing the training dataset size steadily improves the performance of certain methods on the TKGQA task. 

We first apply large pre-trained LM based QA methods on our new dataset. Then we inject KG embeddings, both temporal and non-temporal, into these LMs and observe significant improvement in performance. We also propose a new method, \method{}, that is able to leverage Temporal KG Embeddings to perform TKGQA. In our experiments, \method{} outperforms all baselines. These results suggest that KG embeddings can be effectively used to perform temporal KGQA, although there remains significant scope for improvement when it comes to complex reasoning questions.

\section*{Acknowledgements}

We would like to thank the anonymous reviewers for their constructive feedback, and Pat Verga and William Cohen from Google Research for their insightful comments. We would also like to thank Chitrank Gupta (IIT Bombay) for his help in debugging the source code and dataset. This work is supported in part by a gift from Google Research, India and a Jagadish Bose Fellowship.

\bibliographystyle{acl_natbib}
\bibliography{anthology,acl2021}

\appendix

\section{Appendix}

\subsection{Entities as Experts (EaE)}
\label{sec:eae_appendix}
 The model architecture follows Transformer \citep{vaswani2017attention}  interleaved with an entity memory layer. It has two embedding matrices, for tokens and entities.  It works on the input sequence $x$ as follows.
\begin{equation}
\begin{aligned}
X^0 &= \mathrm{TokenEmbed}(x) \\
X^1 &= \mathrm{Transformer_0}(X^0, \mathrm{num\_layers} = l_0) \\
X^2 &= \mathrm{EntityMemory}(X^1) \\
X^3 &= \mathrm{LayerNorm}(X^2 + X^1) \\
X^4 &= \mathrm{Transformer_1}(X^3, \mathrm{num\_layers} = l_1) \\
X^5 &= \mathrm{TaskSpecificHeads}(X^4)
\end{aligned}
\label{eqn:eae}
\end{equation}
The whole model (transformers, token and entity embeddings, and task-specific heads) is trained end to end using losses for entity linking, mention detection and masked language modeling.

\subsection{EaE for Temporal KGQA}
\dataset{} does not provide a text corpus for training language models.  Therefore, we use BERT \citep{devlin2019bert} for $\mathrm{Transformer_0}$ as well as $\mathrm{TokenEmbed}$ (eqn.~\ref{eqn:eae}). For $\mathrm{EntityMemory}$, we use TComplEx/TimePlex embeddings of entities and timestamps that have been pre-trained using the \dataset{} KG (please refer to Section~\ref{sec:temporal-kg-embeddings} for details on KG embeddings). The modified model is as follows:
\begin{equation}
\begin{aligned}
X^1 &= \mathrm{BERT}(x) \\
X^2 &= \mathrm{EntityTimeEmbedding}(X^1) \\
X^3 &= \mathrm{LayerNorm}(X^2 + X^1) \\
X^4 &= \mathrm{Transformer_1}(X^3, \mathrm{num\_layers} = 6) \\
X^5 &= \mathrm{PredictionHead}(X^4)
\end{aligned}
\label{eqn:eae-add}
\end{equation}
For simplicity, we assume that all grounded entity and time mention spans are marked in the question, i.e., for each token, we know. which entity or timestamp it belongs to (or if it doesn't belong to any).  Thus, for each token $x_i$ in the input~$x$, 
\begin{itemize}
    \item $X^1[i]$ contains the contextual BERT embedding of~$x_i$
    \item For $X^2[i]$ there are 3 cases. 
    \begin{itemize}
        \item $x_i$ is a mention of entity $e$. Then $X^2[i] = \mathcal{E}[e]$.
        \item $x_i$ is a mention of timestamp $t$. Then $X^2[i] = \mathcal{T}[t]$.
        \item $x_i$ is not a mention. Then $X^2[i]$ is the zero vector.
    \end{itemize}
\end{itemize}
$\mathrm{PredictionHead}$ takes the final output from $\mathrm{Transformer_1}$ of the token corresponding to the $\mathrm{[CLS]}$ token of BERT as the predicted answer embedding. This answer embedding is scored against $\mathcal{E.T}$ using dot product to get a score for each possible answer, and $\mathrm{softmax}$ is taken to get answer probabilities. The model is trained on the QA dataset using cross-entropy loss. We will refer to this model as \textbf{T-EaE-add} since we are taking element-wise sum of BERT and entity/time embeddings.

\textbf{T-EaE-replace} Instead of adding entity/time and BERT embeddings, we replace the BERT embeddings with the entity/time embeddings for entity/time mentions. Specifically, before feeding to $\mathrm{Transformer_1}$ in step~4 of eqn.~\ref{eqn:eae-add},
\begin{enumerate}
    \item if $x_i$ is not an entity or time mention, $X^3[i] = \mathrm{BERT}(X^1[i])$
    \item if $x_i$ is an entity or time mention, $X^3[i] = \mathrm{EntityTimeEmbedding}(X^1[i])$
\end{enumerate}
The rest of the model remains the same.

\subsection{Examples}
Tables \ref{tab:example_first} to \ref{tab:example_last} contain some example questions from the validation set of \dataset{}, along with the top 5 predictions of the models we experimented with. T5-3B has a single prediction since it is a text-to-text model.
\begin{table*}[t]

\resizebox{\textwidth}{!}{%
\begin{tabular}{@{}ll@{}}
\toprule
\textbf{Question}           & \textit{Who held the position of Prime Minister of Sweden before 2nd World War}                                                                     \\ \midrule
\textbf{Question Type}      & Before/After                                                                                                                                       \\
\textbf{Gold answer(s)}   & Per Albin Hansson                                                                                                                                   \\ \midrule
\rowcolor[HTML]{EFEFEF} 
\textbf{BERT}               & Emil Stang, Sr., Sigurd Ibsen, Johan Nygaardsvold, Laila Freivalds, J. S. Woodsworth                                                                \\
\textbf{KnowBERT}           & \begin{tabular}[c]{@{}l@{}}Benito Mussolini, Östen Undén, Hans-Dietrich Genscher, Winston Churchill, \\ Lutz Graf Schwerin von Krosigk\end{tabular} \\
\rowcolor[HTML]{EFEFEF} 
\textbf{T5-3B}              & bo osten unden                                                                                                                                      \\
\textbf{EmbedKGQA}   & \textbf{Per Albin Hansson}, Tage Erlander, Carl Gustaf Ekman, Arvid Lindman, Hjalmar Branting                                                                \\
\rowcolor[HTML]{EFEFEF} 
\textbf{T-EaE-add}            & \textbf{Per Albin Hansson}, Manuel Roxas, Arthur Sauvé, Konstantinos Demertzis, Karl Renner                                                                  \\
\textbf{T-EaE-replace}        & \textbf{Per Albin Hansson}, Tage Erlander, Arvid Lindman, Valère Bernard, Vladko Maček                                                                       \\
\rowcolor[HTML]{EFEFEF} 
\textbf{\method{}}        & \textbf{Per Albin Hansson}, Tage Erlander, Arvid Lindman, Carl Gustaf Ekman, Hjalmar Branting                                                                \\ \bottomrule
\end{tabular}%
}
\caption{Before/After reasoning type question.}
\label{tab:example_first}
\end{table*}

\begin{table*}[t]
\centering
\begin{tabular}{@{}ll@{}}
\toprule
\textbf{Question}       & \textit{When did Man on Wire receive Oscar for Best Documentary Feature} \\ \midrule
\textbf{Question Type}  & Simple time                                                             \\
\textbf{Gold answer(s)} & 2008                                                                     \\ \midrule
\rowcolor[HTML]{EFEFEF} 
\textbf{BERT}           & 1995, 1993, 1999, 1991, 1987                                             \\
\textbf{KnowBERT}       & 1993, 1996, 1994, 2006, 1995                                             \\
\rowcolor[HTML]{EFEFEF} 
\textbf{T5-3B}          & 1997                                                                     \\
\textbf{EmbedKGQA}      & 2017, \textbf{2008}, 2016, 2013, 2004                                             \\
\rowcolor[HTML]{EFEFEF} 
\textbf{T-EaE-add}      & \textbf{2008}, 2009, 2005, 1999, 2007                                             \\
\textbf{T-EaE-replace}  & 2009, \textbf{2008}, 2005, 2006, 2007                                             \\
\rowcolor[HTML]{EFEFEF} 
\textbf{\method{}}    & \textbf{2008}, 2007, 2009, 2002, 1945                                             \\ \bottomrule
\end{tabular}
\caption{Simple reasoning question with time answer.}
\label{tab:example2}
\end{table*}

\begin{table*}[t]
\resizebox{\textwidth}{!}{%
\begin{tabular}{@{}ll@{}}
\toprule
\textbf{Question}           & \textit{Who did John Alan Lasseter work with while employed at Pixar}                                                         \\ \midrule
\textbf{Question Type}      & Time join                                                                                                                    \\
\textbf{Gold answer(s)}   & Floyd Norman                                                                                                                  \\ \midrule
\rowcolor[HTML]{EFEFEF} 
\textbf{BERT}               & \begin{tabular}[c]{@{}l@{}}Tim Cook, Eleanor Winsor Leach, David R. Williams, Robert M. Boynton,\\ Jules Steeg\end{tabular}   \\
\textbf{KnowBERT}           & 1994, 1997, Walt Disney Animation Studios, Christiane Kubrick, 1989                                                           \\
\rowcolor[HTML]{EFEFEF} 
\textbf{T5-3B}              & john alan lasseter                                                                                                            \\
\textbf{EmbedKGQA}   & John Lasseter, \textbf{Floyd Norman}, Duncan Marjoribanks, Glen Keane, Theodore Ty                                                     \\
\rowcolor[HTML]{EFEFEF} 
\textbf{T-EaE-add}            & \begin{tabular}[c]{@{}l@{}}John Lasseter, Anne Marie Bardwell, Will Finn, \textbf{Floyd Norman},\\  Rejean Bourdages\end{tabular}      \\
\textbf{T-EaE-replace}        & John Lasseter, Will Finn, \textbf{Floyd Norman}, Nik Ranieri, Ken Duncan                                                               \\
\rowcolor[HTML]{EFEFEF} 
\textbf{\method{}}        & \begin{tabular}[c]{@{}l@{}}John Lasseter, \textbf{Floyd Norman}, Duncan Marjoribanks, David Pruiksma, \\ Theodore Ty\end{tabular}      \\ \bottomrule
\end{tabular}%
}
\caption{Time join type question.}
\label{tab:example3}
\end{table*}
\begin{table*}[t]
\resizebox{\textwidth}{!}{%
\begin{tabular}{@{}ll@{}}
\toprule
\textbf{Question}           & \textit{Where did John Hubley work before working for Industrial Films}                                                                                     \\ \midrule
\textbf{Question Type}      & Before/After                                                                                                                                               \\
\textbf{Gold answer(s)}   & The Walt Disney Studios                                                                                                                                     \\ \midrule
\rowcolor[HTML]{EFEFEF} 
\textbf{BERT}               & \textbf{The Walt Disney Studios}, Warner Bros. Cartoons, Pixar, Microsoft, United States Navy                                                                        \\
\textbf{KnowBERT}           & \begin{tabular}[c]{@{}l@{}}École Polytechnique, Pitié-Salpêtrière Hospital, \textbf{The Walt Disney Studios}, \\ Elisabeth Buddenbrook, Yale University\end{tabular} \\
\rowcolor[HTML]{EFEFEF} 
\textbf{T5-3B}              & london film school                                                                                                                                          \\
\textbf{EmbedKGQA}   & \begin{tabular}[c]{@{}l@{}}\textbf{The Walt Disney Studios}, Collège de France, Warner Bros. Cartoons, \\ University of Naples Federico II, ETH Zurich\end{tabular}  \\
\rowcolor[HTML]{EFEFEF} 
\rowcolor[HTML]{EFEFEF} 
\textbf{T-EaE-add}            & \begin{tabular}[c]{@{}l@{}}\textbf{The Walt Disney Studios}, Fleischer Studios, UPA, Walter Lantz Productions, \\ Wellesley College\end{tabular}                     \\
\textbf{T-EaE-replace}        & \begin{tabular}[c]{@{}l@{}}\textbf{The Walt Disney Studios}, City College of New York, UPA, \\ Yale University, Indiana University\end{tabular}                      \\
\rowcolor[HTML]{EFEFEF} 
\textbf{\method{}}        & \begin{tabular}[c]{@{}l@{}}\textbf{The Walt Disney Studios}, UPA, Saint Petersburg State University, \\ Warner Bros. Cartoons, Collège de France\end{tabular}        \\ \bottomrule
\end{tabular}%
}
\caption{Before/After reasoning type question.}
\label{tab:example4}
\end{table*}

\begin{table*}[t]

\resizebox{\textwidth}{!}{%
\begin{tabular}{@{}ll@{}}
\toprule
\textbf{Question}           & \textit{The last person that Naomi Foner Gyllenhaal was married to was}                                                                                                                 \\ \midrule
\textbf{Question Type}      & First/Last                                                                                                                                                                              \\
\textbf{Gold answer(s)}   & Stephen Gyllenhaal                                                                                                                                                                       \\ \midrule
\rowcolor[HTML]{EFEFEF} 
\textbf{BERT}               & 1928, Jennifer Lash, Stephen Mallory, Martin Landau, Bayerische Verfassungsmedaille in Gold                                                                                              \\
\textbf{KnowBERT}           & Nadia Benois, Eugenia Zukerman, Germany national football team, Talulah Riley, Lola Landau                                                                                               \\
\rowcolor[HTML]{EFEFEF} 
\textbf{T5-3B}              & gyllenhaal                                                                                                                                                                               \\
\textbf{EmbedKGQA}   & \begin{tabular}[c]{@{}l@{}}\textbf{Stephen Gyllenhaal}, Naomi Foner Gyllenhaal, Wolfhard von Boeselager, \\ Heinrich Schweiger, Bruce Paltrow\end{tabular}                                        \\
\rowcolor[HTML]{EFEFEF} 
\textbf{T-EaE-add}            & \textbf{Stephen Gyllenhaal}, Marianne Zoff, Cotter Smith, Douglas Wilder, Gerd Vespermann                                                                                                         \\
\textbf{T-EaE-replace}        & \begin{tabular}[c]{@{}l@{}}\textbf{Stephen Gyllenhaal}, Hetty Broedelet-Henkes, Naomi Foner Gyllenhaal, \\ Miles Copeland, Jr., member of the Chamber of Representatives of Colombia\end{tabular} \\
\rowcolor[HTML]{EFEFEF} 
\textbf{\method{}}        & \begin{tabular}[c]{@{}l@{}}\textbf{Stephen Gyllenhaal}, Antonia Fraser, Bruce Paltrow, \\ Naomi Foner Gyllenhaal, Wolfhard von Boeselager\end{tabular}                                            \\ \bottomrule
\end{tabular}%
}
\caption{First/Last reasoning type question.}
\label{tab:example_last}
\end{table*}



\end{document}